%% file: main.tex
\documentclass[journal]{IEEEtran} 

\usepackage{amsmath}

\usepackage[dvipsnames]{xcolor}

\usepackage{float}

\usepackage{graphicx} 
\graphicspath{{images/}}
\DeclareGraphicsExtensions{.pdf,.PDF,.jpg,.JPG,.jpeg,.JPEG,.png,.PNG}

\usepackage[caption=false]{subfig}

\usepackage{multicol}
\usepackage{multirow}

\usepackage{url}

\usepackage{verbatim}

\usepackage{listings}
\usepackage{acronym}

\newacro{dia}[\textsc{Dia\xspace}]{Document Image Analysis}

\newacro{nln}[\textit{N-light-N}\xspace]{\textit{N-light-N}}

\newacro{scae}[\textsc{SCAE\xspace}]{Stacked Convolutional Auto-Encoder}

\newacro{ae}[\textsc{AE\xspace}]{Auto-Encoder}
\newacroplural{ae}[\textsc{AEs\xspace}]{Auto-Encoders}
\newacro{cae}[\textsc{CAE\xspace}]{Convolutional Auto-Encoder}

\newacro{nn}[\textsc{NN\xspace}]{Neural Network}
\newacroplural{nn}[\textsc{NNs\xspace}]{Neural Networks}
\newacro{ann}[\textsc{ANN\xspace}]{Artificial Neural Network}

\newacro{aenn}[\textsc{AENN\xspace}]{Auto-Encoder Neural Network}

\newacro{cnn}[\textsc{CNN\xspace}]{Convolutional Neural Network}
\newacro{ffcnn}[\textsc{FFCNN\xspace}]{Feed Forward Convolutional Neural Network}

\newacro{dnn}[\textsc{DNN\xspace}]{Deep Neural Network}
\newacroplural{dnn}[\textsc{DNNs\xspace}]{Deep Neural Networks}

\newacro{pca}[\textsc{PCA\xspace}]{Principal Component Analysis}
\newacro{lda}[\textsc{LDA\xspace}]{Linear Discriminant Analysis}

\begin{document}

\title{DeepDIVA: A Highly-Functional Python Framework for Reproducible Experiments}

\author{\textbf{Michele~Alberti}\IEEEauthorrefmark{1}, \and
        \textbf{Vinaychandran~Pondenkandath}\IEEEauthorrefmark{1}, \and
        \thanks{\IEEEauthorrefmark{1} Both authors contributed equally to this work.}
        \textbf{Marcel~W\"ursch}, \and
        \textbf{Rolf~Ingold}, \and
        \textbf{Marcus~Liwicki}\\ 
        \textit{Document Image and Voice Analysis Group (DIVA)} \\
        University of Fribourg
        Switzerland \\
        \{firstname\}.\{lastname\}@unifr.ch
}

\markboth{}%
{Alberti \MakeLowercase{\textit{et al.}}: TODO INSERT TITLE HERE}

\maketitle

\thispagestyle{empty}

\input{sections/abstract}


\input{sections/introduction}

\input{sections/reproducing_experiments.tex}

\input{sections/features.tex}

\input{sections/case_studies.tex}

\input{sections/conclusion}

\input{sections/vision.tex}

\section*{Acknowledgment}
The work presented in this paper has been partially supported by the HisDoc III project funded by the Swiss National Science Foundation with the grant number $205120$\textunderscore$169618$.

\bibliographystyle{IEEEtran}
\bibliography{biblio}

\end{document}

%% file: sections/abstract.tex
\begin{abstract}

We introduce DeepDIVA: an infrastructure designed to enable quick and intuitive setup of reproducible experiments with a large range of useful analysis functionality.
Reproducing scientific results can be a frustrating experience, not only in document image analysis but in machine learning in general.
Using DeepDIVA a researcher can either reproduce a given experiment with a very limited amount of information or share their own experiments with others.
Moreover, the framework offers a large range of functions, such as boilerplate code, keeping track of experiments, hyper-parameter optimization, and visualization of data and results.
To demonstrate the effectiveness of this framework, this paper presents case studies in the area of handwritten document analysis where researchers benefit from the integrated functionality.
DeepDIVA is implemented in Python and uses the deep learning framework PyTorch. 
It is completely open source\footnote{\url{https://github.com/DIVA-DIA/DeepDIVA}}%
, and accessible as Web Service through DIVAServices\footnote{\url{http://divaservices.unifr.ch}}.

\end{abstract}

%% file: sections/introduction.tex
\section{Introduction}
\label{toc:intro}

\begin{figure}[!t]
  \centering
  \subfloat[Comparison of two different training protocols on a watermark classification task. Orange: network initialized with random weights. Pink: network initialized with pre-trained weights]%
  {\includegraphics[width=.47\columnwidth,height=3.6cm]{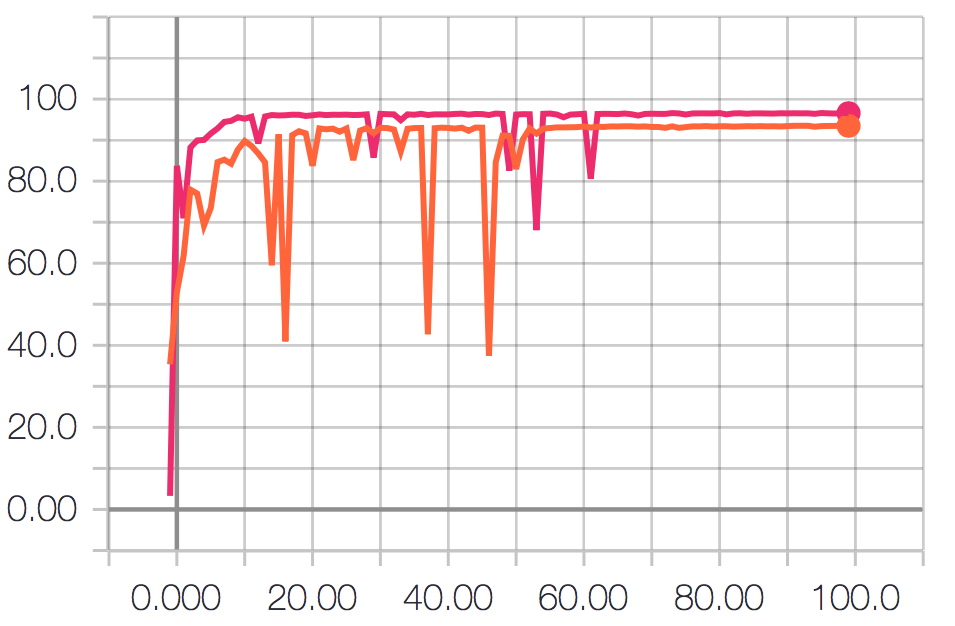}\label{subfig:compareRuns_watermarks}}
  \hfil
  \subfloat[Evaluation of how randomness affects execution by visualizing the aggregated results of multiple runs. Here the shaded area indicates the variance over all runs.]%
  {\includegraphics[width=.47\columnwidth,height=4cm]{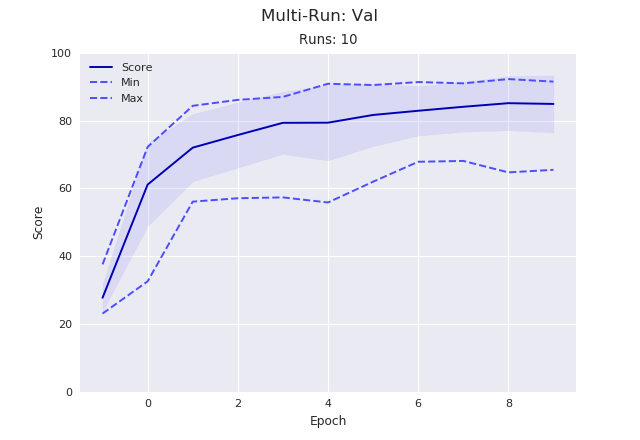}\label{subfig:shadyPlot}}
  
  \vfill
  \subfloat[Confusion Matrix. The darker the color the higher the amount of samples classified as such.]%
  {\includegraphics[width=.47\columnwidth,height=4cm]{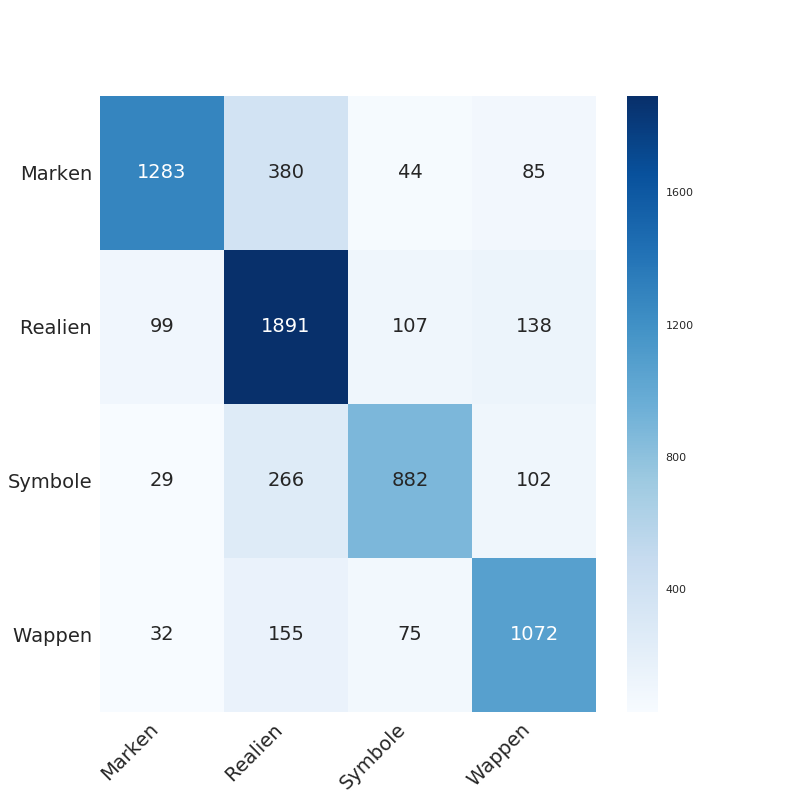}\label{subfig:confusionMatrix}}
  \hfil
  \subfloat[T-Distributed Stochastic Neighbor Embedding (T-SNE) visualization of the watermarks dataset in Tensorboard.]%
  {\includegraphics[width=.47\columnwidth,height=4cm]{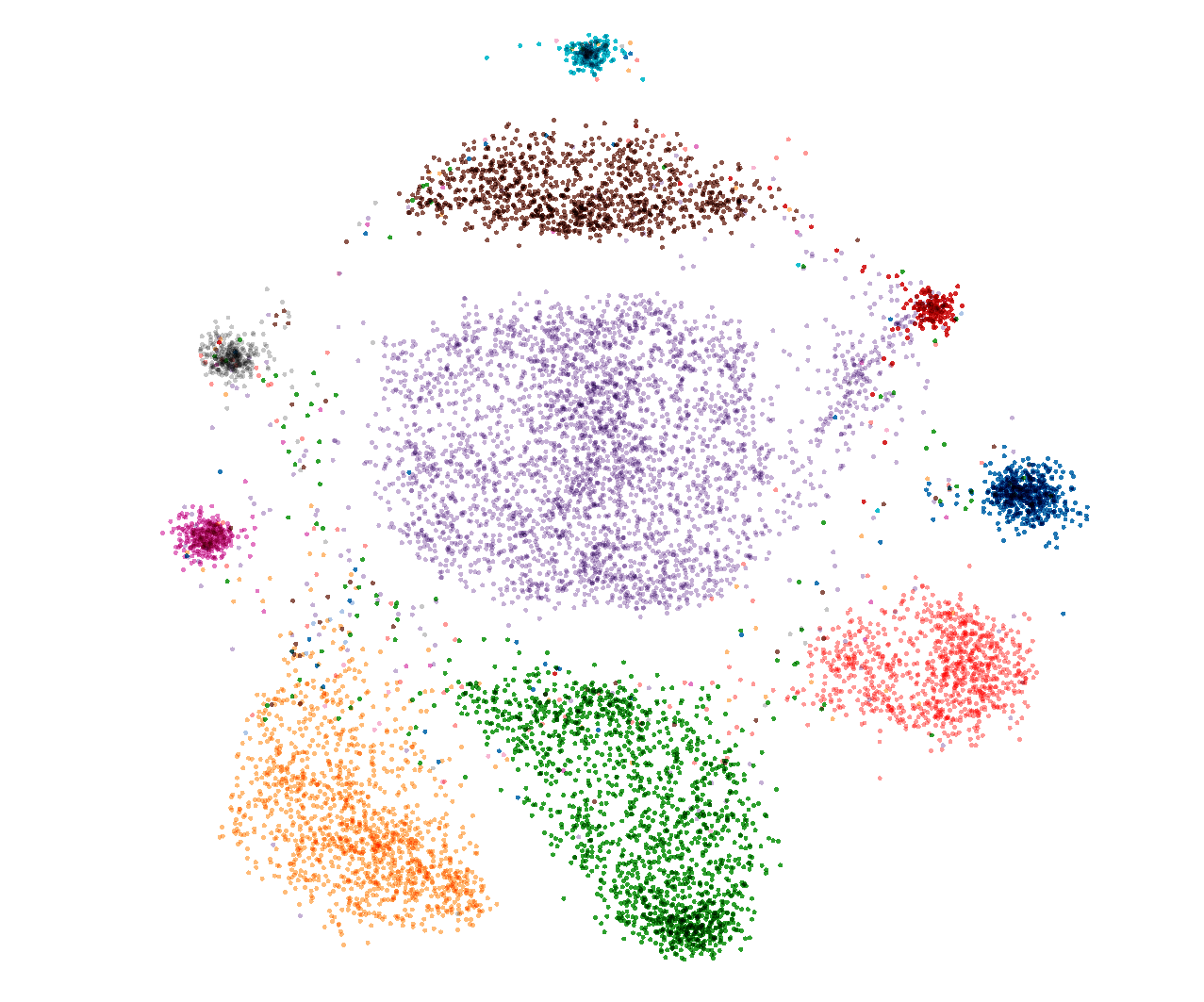}\label{subfig:tsne_watermarks}}
  \caption{Example of different visualizations produced automatically by DeepDIVA.}
  \label{fig:visualization_produced}
\end{figure}

\begin{figure}[!t]
  \vfill
  \subfloat[Output decisions at epoch 1]%
  {\includegraphics[width=.47\columnwidth,height=4cm]{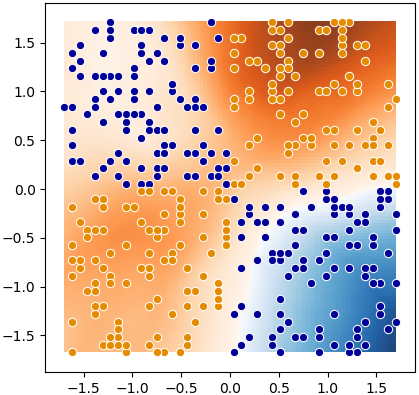}\label{subfig:e0}}
  \hfil
  \subfloat[Output decisions at epoch 10]%
  {\includegraphics[width=.47\columnwidth,height=4cm]{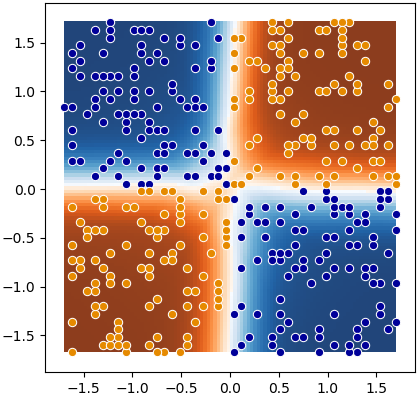}\label{subfig:e10}}
  \caption{Visualization of the decision boundaries of a network learning the continuous XOR problem. Color intensity represent the confidence of the network.}
  \label{fig:2D_visualizations}
\end{figure}

An important topic, not only in handwritten historical document image analysis, but in machine learning in general, is the reproducibility of scientific results. 
Often, we have scientific publications in our hands showing results difficult to reproduce.
This phenomenon, known as the reproducibility crisis, is a methodological phenomenon in which many prominent studies and papers have been found to be difficult or impossible to reproduce. 
The replication crisis has been widely studied in other fields~\cite{john2012measuring, begley2013reproducibility, camerer2016evaluating} and is getting more attention in the machine learning community, recently~\cite{hutson2018artificial, olorisade2017reproducibility}.
This even led to prompting a replication challenge~\cite{iclr_2018_reproducibility_challenge} for reproducing the results of prominent papers. 

Traditionally, scientific experiments are to be described in a way that allows anyone to replicate them. 
However, especially in computer-science, there are many complications which makes complete reproducibility particularly difficult to achieve. 
This can be attributed to differences in software versions, specific dependencies, or even hardware variations.
Furthermore, specific details are sometimes confidential and the authors are not allowed to include all hints leading to successful reproduction.
In many cases, details are just omitted, because they are thought to be obvious or they would require too much space which is rare due to the page limit.
Especially in such cases, publishing of the source code or even providing the whole experimental set-up would be beneficial.

To address many of these issues, we develop a framework to help researchers in both, saving time on common research-oriented software problems and sharing experiments by making them easier to reproduce. 
Our framework, DeepDIVA, is originally designed to perform deep-learning tasks on historical document images.
However, thanks to it's flexible and modular design it can be easily extended to work in other fields or new tasks.
We hope to engage the community in building a shared tool which would eventually foster experiment reproducibility and higher quality of code, benefiting newcomers in the field as well as experts.

\subsection*{Main Contribution}
\label{toc:Main Contribution}

In this paper, we introduce DeepDIVA: an open-source Python deep-learning framework which aims at easing the life of researchers by providing a variety of useful features out-of-the-box.
DeepDIVA is designed to enable experiment reproduction with minimal effort.
The framework integrates popular tools such as SigOpt~\cite{sigopt} for hyper-parameter optimization and TensorBoard\footnote{\url{https://www.tensorflow.org/programmers_guide/summaries_and_tensorboard}} for aggregating all visualization produced (see Fig.~\ref{fig:visualization_produced}).
Additionally, we provides boilerplate code for standard deep-learning experiments and especially for experiments on Handwritten Documentents. 
It also offers the possibility to be run as a Web Service on D\textsc{iva}Services \cite{Wursch2018}. 

\subsection{Related Work}
\label{toc:Related Work}

DeepDIVA shares the spirit and motivation with several existing frameworks and tools.
Neptune~\cite{neptune2017} is a commercial solution that offers a variety of features (similar to DeepDIVA) and also has extended support for cloud computing.
Comet~\cite{comet2017} lets you track code, experiments, and results and works with most machine learning libraries.
CodaLab~\cite{codaLab2014} provides experiments management trough a proprietary web interface and allows to re-run other peoples experiments.
Data Version Control~\cite{dataVersionControl2017} is an open-source tool that manages code and data together in a simple form of Git-like commands.
Polyaxon~\cite{polyaxon2018} is an open-source platform for building, training, and monitoring large-scale deep learning applications.
It allows for deployment to different solutions such as data centers or cloud providers.
OpenML~\cite{vanschoren2013openml} is a tool which focuses on making datasets, tasks, workflows as well as results accessible for free. 
Many other tools, such as Sumatra~\cite{davison2012automated}, Sacred~\cite{greff2017sacred}, CDE~\cite{howe2012cde}, FGBLab~\cite{arulkumaranfglab} and ReproZip~\cite{chirigati2016reprozip}, collect and store much different information about sources, dependencies, configurations, the file used, host information and even system calls, with the aim of supporting reproducible research.
These tools collect and store different information about sources, dependencies, configurations, the file used, host information and even system calls, with the aim of supporting reproducible research.
However, most or these tools strive being independent of the choice of machine learning libraries.
While this leads to more flexibility and a broader scope, it limits the functionality (e.g., no  Graphics Processing Units (GPU) support, no prototype implementations, no visualizations) and makes the tools heavy.
DeepDIVA relies on a working Python environment and specific settings. 
This makes it lightweight and easy to cope with GPU hardware. 
DeepDIVA provides basic implementations for common tasks and has a wide-ranging support for visualizations.
Additionally, it addresses the situation where an algorithm has non-deterministic behavior (random initialization for example).

%% file: sections/reproducing_experiments.tex
\section{Reproducing Experiments}
\label{toc:reproducing_experiments}

When reproducing experiments of others, the main issue is the code unavailability followed by the --- sometimes prohibitive --- amount of time necessary to make it work.
Unlike other tools that offer reproducibility by trying to cope with heterogeneous scenarios, we provide an easy-to-use environment that any researcher can easily adopt.

Our main paradigm for making experiments reproducible for anyone is an enforced version control.
While it is a standard in modern software development for enabling tracing of the development progress, researchers often have the practice of \{un\}commenting lines of code and re-run the experiment again, making it almost impossible to match results to a code state.
DeepDIVA uses two alternative ways to ensure that every single experiment can be reproduced.
By default, the framework checks if all code is committed before running an experiment. 
Git\footnote{\url{https://git-scm.com/}} creates a unique identifier for each commit and this is also the information that can be shared with others to reproduce the experiment.
This way is the only way to create results seamlessly reproducible to other researchers.
To ensure that the repository name, version, and parameters don't get lost, all information is saved together in the results file.
Note that always creating a version would be very cumbersome during development. Therefore, this behavior can be suppressed.
In that case, DeepDIVA creates an actual copy of the code in the output folder at each run.

In order to reproduce an experiment one needs to know the following three pieces of information: Git repository URL, Git commit identifier, and the list of parameters to use.
Reproducing an experiment is easily performed by cloning the Git repository\footnote{For details on how to setup the environment see Section~\ref{toc:Deep-learning out-of-the-box} or the README file on DeepDIVA repository directly.} at the specific commit identifier and run the entry point of the framework with the list of arguments as command line parameters. 

Even when all code and parameters are provided,  certain pitfalls exist that need to be taken care of.
The most common is seeding pseudo-random generators.
This is a nontrivial issue when working with both CPU and/or multiple GPUs.
In our framework, if the seed is not specified by the user, we select a random value, log it, then use it for seeding Python, NumPy, and PyTorch pseudo-random generators.
However, this is not enough for ensuring the \textit{exact} same values for each run due to the non-deterministic nature of the optimized NVIDIA Compute Unified Device Architecture (CUDA) Deep Neural Network library (CuDNN) kernels that PyTorch uses, therefore we allow the option to disable CuDNN for increased reproducibility.\footnote{Enabling \texttt{ torch.backends.cudnn.deterministic} showed to be not working correctly in our tests}.

%% file: sections/features.tex
\section{Features}
\label{toc:features}

Apart from reproducibility, DeepDIVA has additional features  helping researchers in several common scenarios.
This section briefly describes the most important features and how they can be beneficial. 

\subsection{Deep-learning Out-of-the-Box}
\label{toc:Deep-learning out-of-the-box}
One of the greatest strength of the framework is its simple set up.
All one needs to do is clone the repository from Github and to execute the setup script. 
This script sets up a local \textit{Conda}\footnote{Conda is an open source package management system and environment management system that runs on Windows, macOS, and Linux. See \url{https://conda.io/docs/}} environment with all the required packages. 
For GPU compatibility, a system with the appropriate GPU drivers must be used for the set-up.

At the moment of writing, DeepDIVA features implementations of boilerplate for three common use cases: image classification, image-patch matching and various scenarios with bi-dimensional data.
As different researchers have different needs, hence the framework is designed to be fully customizable with minimum effort.
In practice, this is achieved by having a high modularity between the different components such as data preparation, model definition, train, validation and test routines.

This is particularly useful in deep-learning, where it is quite often the case that implementing the boilerplate constitutes the bulk part of programming workload.
For example, one only needs to swap out a default component and replace it with their own implementation of it, i.e., a new model architecture or a new training routine. 

\subsection{Automatic Hyper-parameter Optimization}
\begin{figure}[t]
    \centering
    \includegraphics[width=\columnwidth, height=5cm]{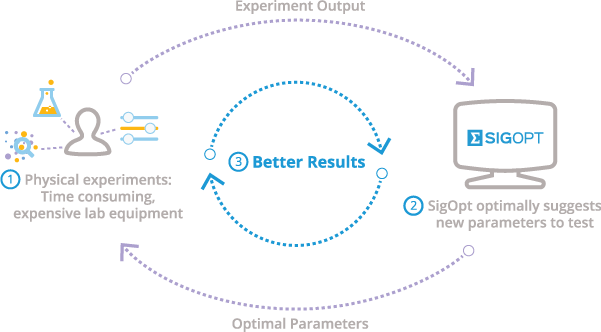}
    \caption{Example of hyper-parameter optimization process with SigOpt. In DeepDIVA this happens in a completely automated fashion where a list of input parameters is optimized thank to an iterative exchange of optimal configurations suggested by SigOpt tool.}
    \label{fig:sigopt}
\end{figure}

Nowadays virtually all deep-learning experiments require a certain amount of hyper-parameters optimization. 
This is a tedious and time-consuming procedure which often does not require a real interaction from the researcher. 
A lot of research has been done to find an efficient way to optimize parameters other than random or grid search.
Therefore, we integrate Bayesian hyper-parameter optimization into the framework using SigOpt~\cite{sigopt}.
Instead of exhaustively testing every option, SigOpt adaptively suggests experimental configurations (see Fig.~\ref{fig:sigopt}) such to maximize/minimize your chosen metrics, e.g., accuracy, loss.

\subsection{Data Visualization}
\begin{figure}[!t]
  \centering
  \includegraphics[width=\columnwidth, height=7 cm]{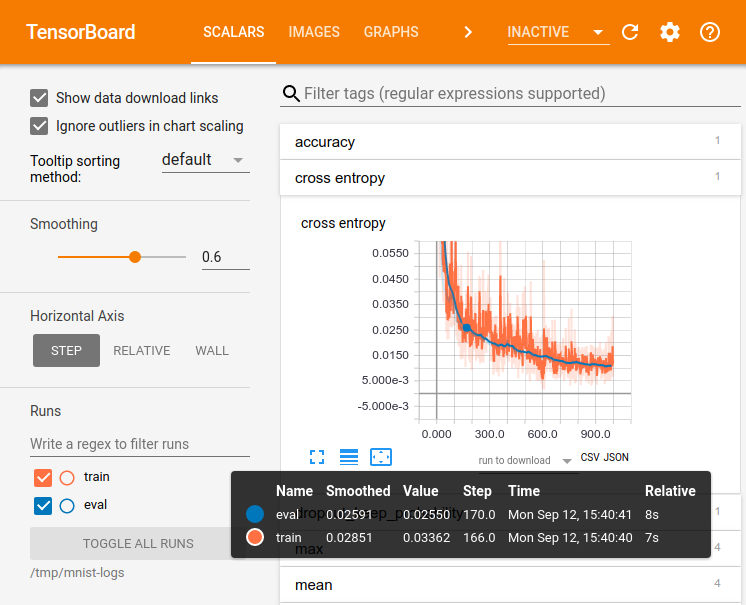}
  \caption{Example of the Tensorboard web interface. Figure from TensorFlow\textsuperscript{TM}}
  \label{fig:tensorboard}
\end{figure}

One of the most common recurring tasks of a researcher is visualizing the data produced by their experiments. 
Visualizations are very helpful not only for understanding, debugging, and optimizing programs but also to make it easier for a human to make sense out of a huge amount of numbers, e.g., by showing trends and behavior of a neural network.
There are several tools that can produce such visualizations, but often these tools are either hard to integrate or cover only a specific kind of visualization (e.g., only plots) and therefore one must resort using multiple ones.
In DeepDIVA, we integrate Tensorboard (see Fig.~\ref{fig:tensorboard}) as it natively supports a wide range of visualizations and it provides easy means to display data produced by any other source. 
This way all the visualizations produced by the framework are aggregated into a single web interface where they can be viewed with real-time updates.
This is particularly helpful when the run time of a program is very long - having early results can save a lot of time when optimizing.

\subsection{Comparing Methods}
Comparing the performance of two or more algorithms --- or multiple runs of the same one with different configurations -- is another task that a researcher often has to deal with.
Tensorboard offers a solution to compare two or more executions (see Fig.~\ref{subfig:compareRuns_watermarks}) as the plots are generated dynamically when selected in the web interface. 
In order to visualize and understand the influence of randomness on an experiment, DeepDIVA has a multi-run flag which automatically runs the same experiment a specified number of times and the aggregates the result into a plot (see Fig.~\ref{subfig:shadyPlot}).
Finally, a common thing to do while evaluating (e.g., a classifier), is to produce the confusion matrix.
DeepDIVA automatically generates one (see Fig.~\ref{subfig:confusionMatrix}) and it is visible in Tensorboard as soon as it is created.

\subsection{Internal Inspection}
When analyzing neural networks looking at the network from outside is often not enough. 
A wide array of tools and technologies have been developed to allow the inspection of neural networks as they run or to visualize their internal status.
DeepDIVA offers the possibility to visualize how the histograms of the weight distributions for each layer evolve over time.
This is useful to spot anomalies, especially when working with various neural network initializations.
We plan to offer more of this kind of insights in the future.

\subsection{Working on 2D Data}

A common practice when developing a new instrument is starting on a controlled toy-problem and then gradually transition into more complex scenarios. 
This is sometimes challenging to achieve in deep-learning due to the high dimensional nature of the data.
To help in this regard, DeepDIVA offers a whole suite of tools that allows one to train, evaluate and visualize performance on different types of bi-dimensional data.
The advantage of using 2D points is that the visualization of the network conforms to the actual data.
An example of such a visualization is given in Fig.~\ref{fig:2D_visualizations} where the behavior of a network learning a continuous XOR problem is shown.
It is not always possible to use such a simple setting as toy-problem, but we believe this is particularly useful when making fundamental research on new technologies rather than trying to push the boundaries of a given model/architecture. 
To allow full customization of these toy-problems there is a script in DeepDIVA which allows a user to create their own custom 2D datasets. 

\subsection{New Dataset Support }
When working on a new dataset it can be the case that a set of operations needs to be performed before a deep-learning model can perform on it.
One of these operations is creating a custom \texttt{dataset} class and custom \texttt{dataloaders}\footnote{This naming is specific to PyTorch but the concept generalizes to other frameworks as well.}. 
In DeepDIVA it is only required to provide the path to the folder where the dataset is located on the file system and the framework will take care of the modifications such as re-sizing the input to the fit the network specifications\footnote{This is an annoying problem e.g. when using a CNN on MNIST and then switching to CIFAR.}.
Current limitations are that the datasets must be of images (no audio, video or graphs yet) and must be organized in a common standard folder format.
Another of these preparation tasks for with DeepDIVA offers automated support is computing the mean and standard deviation of the dataset. 
This is widely done in deep-learning to normalize the dataset before training a model and can be non-trivial if the datasets are larger than the memory of the system - which is often the case when working on historical documents.
We additionally provide a script for partitioning the training data into train and validation subsets.

\subsection{DeepDIVA as a Web Service}
We provide access to DeepDIVA through a Web Service on D\textsc{iva}Services\cite{Wursch2018}.
This Web Service is able to reproduce results as described in Section~\ref{toc:reproducing_experiments}, thus allowing everyone to replicate experiments without the need for any local installations.
To facilitate this, we built a Docker image, that contains a DeepDIVA base installation as well as a script that executes the necessary steps to perform the replication of an experiment. 
This Docker image can be used by others and is hosted on Docker Hub\footnote{see: \url{https://hub.docker.com/u/divaservices/}, this URL is a placeholder and will link to the correct Image in the Camera Ready version}.
We also provide a bash script that can be used to replicate the second case study described in Section \ref{toc:case:study}. This script is available online\footnote{see: \url{https://github.com/DIVA-DIA/DIVAgain/tree/master/ICFHR2018_DeepDIVA_WriterIdentification}}.
For the future, we want to provide the full set of features of DeepDIVA through this Web Service.

%% file: sections/case_studies.tex
\section{Case Studies for Handwritten Documents}
\label{toc:case:study}

In this Section, we present two use cases to demonstrate the usefulness of the framework for handwriting related tasks.

\subsection{ResNet for Watermark Recognition}

A common scenario in many computer vision tasks is classification of images. 
In the context of handwritten historical document image analysis, tasks such writer identification, image/line segmentation, script recognition can be treated as image classification tasks. 

\begin{figure}[!t]
  \centering
  \subfloat[]%
  {\includegraphics[height=3.7cm]%
  {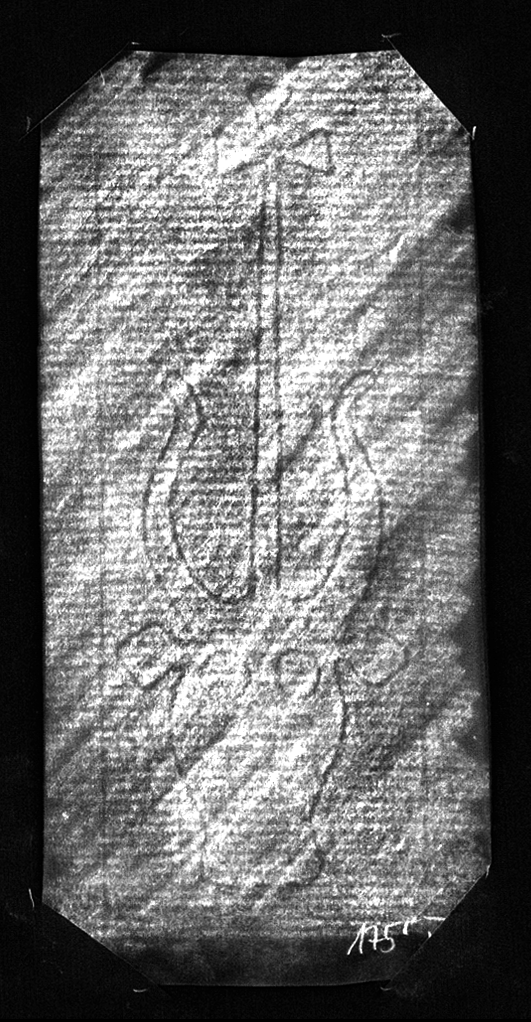}\label{subfig:watermarks_berge1}}
  \hfil
  \subfloat[]%
  {\includegraphics[height=3.7cm]%
  {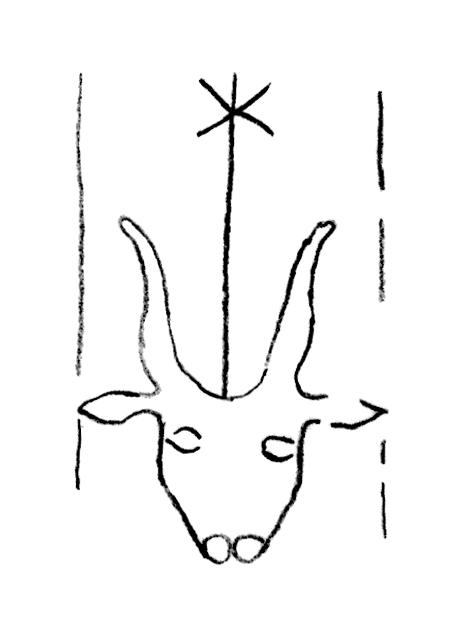}\label{subfig:watermarks_berge2}}
  
  \caption{Sample from the watermarks dataset (a) juxtaposed with a correct (b) classification results from the network. Notice how the way they are depicted is significantly different.}
  \label{fig:watermarks}
\end{figure}

As a use case, we demonstrate how to perform watermark recognition on a dataset provided by the watermark database Wasserzeichen Informationssystem\footnote{ https://www.wasserzeichen-online.de/wzis/struktur.php}. 
The dataset contains images created with different methods of reproduction, such as hand tracing, rubbing, radiography and thermography (see Fig.~\ref{fig:watermarks}). 
We tackle this classification problem using a Convolutional Neural Network (CNN).

Support for classification tasks is provided in DeepDIVA as a basic template, therefore, the only pre-processing necessary for the dataset is to arrange the dataset in a commonly used folder format. 
This, however, needs to be performed on a per-dataset basis, as the source format for each dataset can differ significantly. 
Once the dataset has been transformed into the specified format, you can begin training your desired neural network. 

To train your network, the only required arguments are the \textit{dataset path},\textit{experiment name} and \textit{model}.
Various other parameters such as \textit{optimizer}, \textit{learning rate}, \textit{number of training epochs}, \textit{batch size} and many others can be specified if required.

Several models are provided in a model directory inside of DeepDIVA, and it is easy to add a required model to our framework (copy the defined model source code into the model directory and add the new model as an import in the \texttt{init.py} file in the directory).
The data does not need to be resized to the dimensions required by the dataset, as data transformations are built-in. 
DeepDIVA is fully compatible with multi-GPU support and also supports multiple workers for loading and pre-processing data. 

For this particular task, we train an 18-layer Residual CNN \cite{he2016deep}  pre-trained on the ImageNet dataset (ImageNet pre-training has been shown to be useful for several document image tasks~\cite{afzal2015deepdocclassifier}. 
We train our network using the Stochastic Gradient Descent optimizer with a learning rate of \texttt{0.01} for \texttt{100} epochs. 

To compare the effect of the pre-training, we initialize another instance of the same model with random weights for the neural network.
Both networks are seeded with the same random value and use the same training protocol to prevent any variations in the order in which the data is shown to the network. 
A comparison of their classification accuracy on the validation set can be seen in Fig.~\ref{subfig:compareRuns_watermarks} and the embeddings produced by the final layer are visualized in Fig.~\ref{subfig:tsne_watermarks}. The final performance of the pre-trained network on the test set is \texttt{96.4\%} accuracy. 


Running DeepDIVA for a classification task can be done with this command:

\begin{lstlisting}[breaklines=true]
$ python template/RunMe.py --dataset-folder /dataset/watermarks_wzis/ --model-name resnet18 --experiment-name watermark_classification --epochs 100 --seed 42 --lr 0.01
\end{lstlisting}

Note that more details on these experiments and how to tackle the cross-depiction problem can be found explained in detail in the original work~\cite{pondenkandath2017watermarks}.

\subsection{Triplet Networks for Writer Identification}
\label{toc:writer:classification}

Here we identify the authorship of a document, based only on the document images. We use the dataset provided by the ICDAR2017 Competition on Historical Document Writer Identification~\cite{fiel2017}
This is a particularly challenging task, as the train and test sets do not have the same writers. 
To accomplish this style of identification, we train a CNN using the triplet margin loss metric~\cite{wang2014learning} to embed document images in a latent space such that documents belonging to the same writer are close to each other. 

As this is not a standard classification task, we make some modifications to the framework. 
However, since DeepDIVA is designed to facilitate easy modifications, it is fairly easy to add in your own code and still benefit from all its features. 
To incorporate the triplet network into DeepDIVA, we create a new template that inherits from the standard classification template. 
This allows us to only override methods that require changes, e.g., the training method requires a triplet of input and uses the triplet margin loss\footnote{Additional changes can be seen in an example of the code available at \url{https://github.com/DIVA-DIA/DIVAgain/tree/master/ICFHR2018_DeepDIVA_WriterIdentification}}. 

\begin{figure*}[!t]
    \centering
    \subfloat[Query image]%
    {\includegraphics[width=4cm,height=3.7cm]%
    {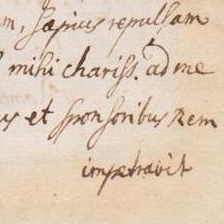}\label{subfig:writer_ident_1}}
    \hfil
    \subfloat[Result One]%
    {\includegraphics[width=4cm,height=3.7cm]%
    {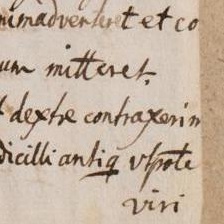}\label{subfig:writer_ident_2}}
    \hfil
    \subfloat[Result Two]%
    {\includegraphics[width=4cm,height=3.7cm]%
    {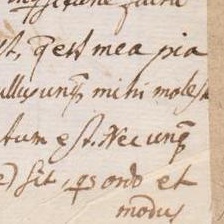}\label{subfig:writer_ident_3}}
    \hfil
    \subfloat[Result Three]%
    {\includegraphics[width=4cm,height=3.7cm]%
    {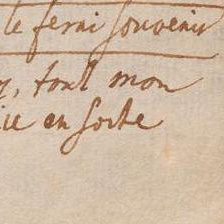}\label{subfig:writer_ident_4}}
      
    \caption{
       For the query image (a), results (b) and (c) belong to the same writer and (d) belongs to a different writer.
    }
    \label{fig:writer_ident}
\end{figure*}

Similarly to the previous task, we use an ImageNet pre-trained CNN for this task. Inputs for the network consist of cropped sub-images from the document images. 
Three such images are input into the network and then the triplet loss is computed based on the distance between the embedding vectors in the latent space. 
We train this network for \texttt{10} epochs.
Since in this dataset there are only 2 positive matches for each query image we measured  \texttt{Precision@1} and \texttt{Precision@2} achieving scores \texttt{0.61}, \texttt{0.69} respectively. 
An example query and returned results for our trained system\cite{pondenkandath2017exploiting} can be seen in Fig.~\ref{fig:writer_ident}.

%% file: sections/conclusion.tex
\section{Conclusion and Future Work}
\label{toc:conclu}

We introduce DeepDIVA: an open-source Python deep-learning framework designed to enable quick and intuitive setup of reproducible experiments with a large range of useful analysis functionality.
We demonstrate that it increases reproducibility and supplies an easy way to share research among peers.
Moreover, we show how researchers can benefit from its features in their daily life experiments thus saving time while focusing on the analysis.

%% file: sections/vision.tex

In the near future we will include in DeepDIVA soon will include the possibility to initialize a network with advanced techniques, such as Principal Component Analysis~\cite{seuret2017pca} or Linear Discriminant Analysis~\cite{alberti2017lda}. 
Additionally, we plan adding more visualization tools, such as activation maps of intermediate layers, heat-maps, and loss landscapes~\cite{li2017}.